\documentclass{article}

\usepackage[nonatbib,preprint]{neurips_2022}

\usepackage[utf8]{inputenc} 
\usepackage[T1]{fontenc}    
\usepackage{hyperref}       
\usepackage{url}            
\usepackage{booktabs}       
\usepackage{amsfonts}       
\usepackage{nicefrac}       
\usepackage{microtype}      
\usepackage{xcolor}         

\usepackage{amsmath}
\usepackage{amssymb}
\usepackage{mathtools}
\usepackage{amsthm}

\usepackage{multirow}
\RequirePackage{algorithm}
\RequirePackage{algpseudocode}

\algblock{ParFor}{EndParFor}
\algnewcommand\algorithmicparfor{\textbf{parfor}}
\algnewcommand\algorithmicpardo{\textbf{do}}
\algnewcommand\algorithmicendparfor{\textbf{end\ parfor}}
\algrenewtext{ParFor}[1]{\algorithmicparfor\ #1\ \algorithmicpardo}
\algrenewtext{EndParFor}{\algorithmicendparfor}

\theoremstyle{plain}

\theoremstyle{definition}

\theoremstyle{remark}

\newcommand{\CC}{\mathcal{C}}
\newcommand{\abs}[1]{\left\lvert {#1} \right\rvert}

\usepackage[textsize=tiny]{todonotes}

\newcommand{\red}[1]{\textcolor{black}{#1}}

\newcommand{\black}[1]{\textcolor{black}{#1}}

\newcommand{\vcenteredinclude}[1]{
    \includegraphics[height=2.5\fontcharht\font`\B]{#1}
}

\newcommand{\emoji}{%
    \begingroup\normalfont
    \hspace*{-1\fontcharht\font`\B}
    \raisebox{-0.7ex}{\vcenteredinclude{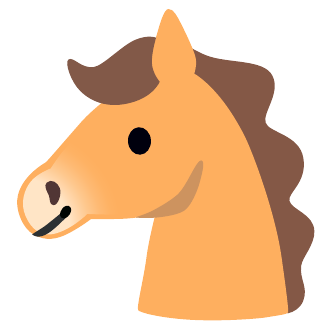}}
    \hspace*{-1\fontcharht\font`\B}
    \endgroup
}

\usepackage{natbib}
\usepackage{bigdelim}

\title{\vspace{-10px}STable \emoji{} Table Generation Framework for~Encoder-Decoder Models}

\author{%
  Michał Pietruszka$^*$
  \quad
  Michał Turski$^*$
  \quad
  Łukasz Borchmann$^*$
  \quad
  Tomasz Dwojak
  \And
  Gabriela Pałka
  \quad
  Karolina Szyndler
  \quad
  Dawid Jurkiewicz
  \quad
  Łukasz Garncarek\vspace{5mm}\\
  Snowflake\\
  \texttt{name.surname@snowflake.com}
}

\begin{document}

\maketitle

\begin{abstract}
\vspace{-2.5mm}\looseness=-1 The output structure of database-like tables, consisting of values structured in horizontal rows and vertical columns identifiable by name, can cover a wide range of NLP tasks. Following this constatation, we propose a framework for text-to-table neural models applicable to problems such as extraction of line items, joint entity and relation extraction, or knowledge base population. The permutation-based decoder of our proposal is a generalized sequential method that comprehends information from all cells in the table. The training maximizes the expected log-likelihood for a table's content across all random permutations of the factorization order. During the content inference, we exploit the model's ability to generate cells in any order by searching over possible orderings to maximize the model's confidence and avoid substantial error accumulation, which other sequential models are prone to. Experiments demonstrate a high practical value of the framework, which establishes state-of-the-art results on several challenging datasets, outperforming previous solutions by up to 15\%.
\end{abstract}

\section{Introduction}\label{sec:introduction}

\begin{figure}[h]
    \centering
    \includegraphics[width=0.9\linewidth]{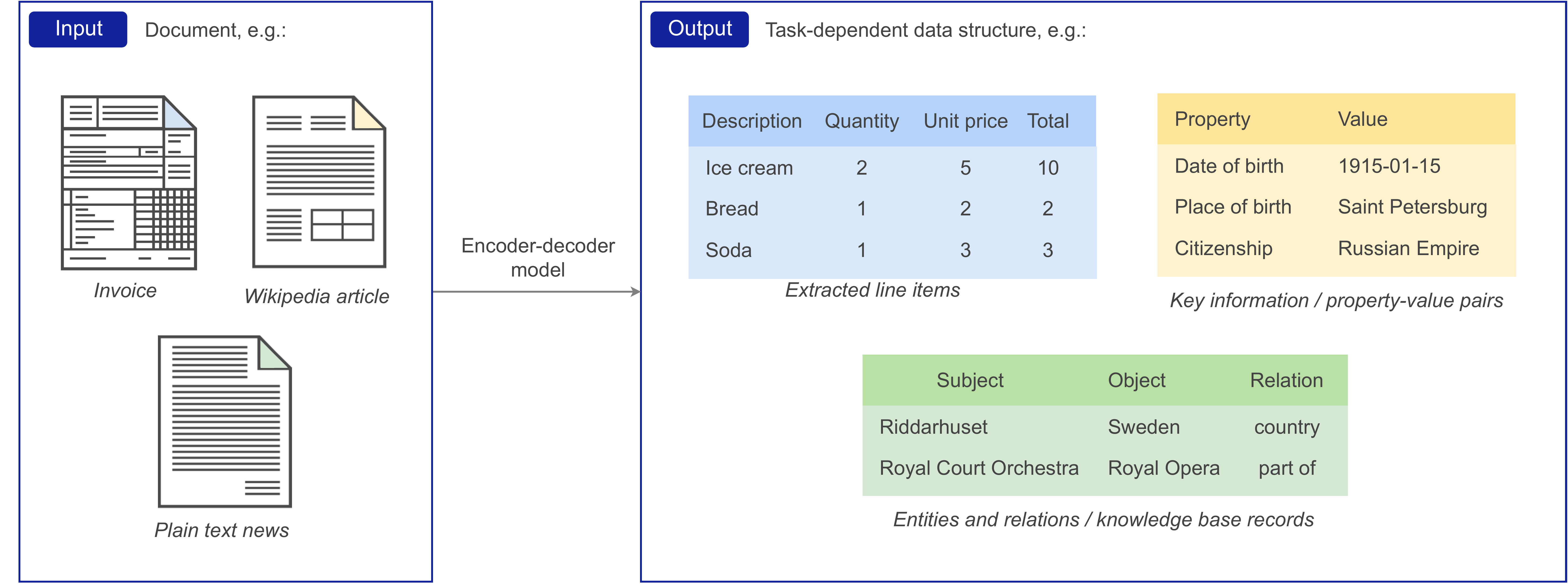}
    \caption{Variety of problems unified as conditioned table generation. 
    }
    \label{fig:hero}
\end{figure}

It has been previously shown that encoder-decoder models are capable of unifying a variety of problems involving natural language. In this setting, unification is achieved by casting different tasks as Question Answering with a plain-text answer, i.e., assuming the text-to-text \citep{pmlr-v48-kumar16,2020t5,DBLP:journals/corr/abs-1806-08730,khashabi2020unifiedqa} or document-to-text scenario \citep{powalski2021tilt}. We argue that the restriction of output type to raw text is suboptimal for the plethora of NLP problems and propose a decoder architecture able to infer \textit{aggregate} data types such as a list of ordered tuples or a database-like table (see Figure~\ref{fig:hero}).

\looseness=-1 Though the encoder-decoder architecture was formerly used to infer lists \citep{powalski2021tilt}, named tuples \citep{dwojak-etal-2020-dataset}, or even more complex structures \citep{DBLP:journals/corr/abs-2105-07510}, it was often achieved in an autoregressive manner, without any architectural changes. For example, a model intended for the generation of \textit{unstructured} text in natural language was used to infer an output with formal \textit{structure}. In contrast, we exploit regularities and relationships within the output data and employ a grammar-constrained decoding process. \black{Consequently, incorrect tables cannot be generated. Part of these rules is explicit (e.g., we overwrite logits, so it is impossible to emit particular tokens such as the end-of-cell when no cell is opened), whereas part of the rules results implicitly from the algorithm.}
%

Specifically, we focus on the text-to-table inference with applications to problems such as extraction of line items, key information extraction of multiple properties, joint entity and relation extraction, or knowledge base population. Tables as we understand them are equivalent to database tables and defined as a set of values structured in horizontal rows and vertical columns identifiable by name. 

\subsection{Motivation}
\label{sec:motivation}
The problem we address with a new model architecture occurs in the context of information extraction from both plain text and richly formatted inputs. From receipts and invoices, through paycheck stubs and insurance loss run reports, to scientific articles, real-world documents contain explicitly or implicitly tabular data to be extracted.

\begin{figure}
    \centering
    \includegraphics[width=0.8\linewidth]{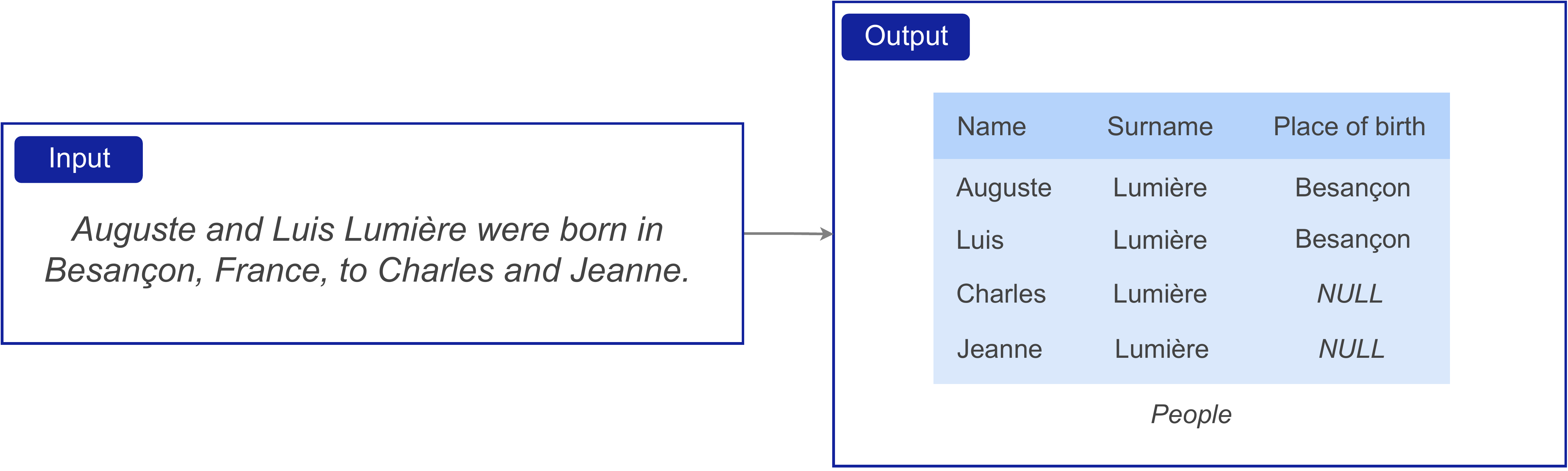}
    \caption{Example of text-to-table generation given plain text input. Concurrent extraction and grouping of the detected entities simplifies the process and may mitigate error accumulation.}
    \label{fig:example}
\end{figure}

These are not necessarily represented as a table \textit{per se} within the input document, e.g., the currency name on the invoice or policy number on the loss run can be mentioned once and be related to all the line items within. In other cases, the evidence one intends to comprehend and represent as a table may be available in free-text only, as can be found in problems of joint entity and relation extraction (see Figure~\ref{fig:hero}-\ref{fig:example}). Finally, the data may require some postprocessing, such as the normalization of dates, before returning them to the end-user.

Moreover, it has been shown that parallel extraction of several property-value pairs can lead to better results than the extraction of these values one-by-one \citep{dwojak-etal-2020-dataset}. We intend to achieve such emergent properties under the same decoding framework.

\looseness=-1 Significantly, the advantage the encoder-decoder framework has is that it can cover problems mentioned above in one end-to-end trainable process, thus simplifying the pipeline and reducing the accumulation of errors along the way. At the same time, since extracted data is already in the form the end user requires, one is able to use it directly for downstream application without further processing steps.
%


Admittedly, models based on the transformer encoder-decoder or decoder achieve remarkable results in generating complex, formalized outputs, such as computer programs or JSON files \citep{DBLP:journals/corr/abs-2107-03374,DBLP:journals/corr/abs-2105-07510}. 
Nevertheless, we hypothesize that architectural changes leading to the \textit{explicit} modeling of structured data can outperform the said \textit{implicit} decoding that models long-range syntax dependencies sequentially and does not guarantee the formal validity of produced outputs.

\subsection{Contribution and Related Works}\label{sec:contributions}

The specific contribution of this work includes (1) equipping transformer models with permutation-based decoder training to allow comprehending complex, role-dependent relationships in a series of similar objects we represent as a table, and (2) a sequential, grammar-constrained decoding mechanism which generates table content cell-by-cell, in a dynamic, data-dependent order.
The novelty of our approach can be better understood in the context of related works.

\looseness=-1 \textbf{Decoding of data structures.} A few authors attempted the problem of table generation in the encoder-decoder framework. \citet{10.1007/978-3-030-58589-1_34} proposed a table recognition model consuming input images and decoupled the problem into unconstrained table and cell content generation. In comparison, (1) we use a single constrained decoder comprehending both table structure and its content; (2) we tackle problems of text-to-table inference where the presence of a table at the model input is optional. {\color{black}Recently, \citet{DBLP:journals/corr/abs-2109-02707} introduced a model relying on constrained decoding of table and tabular embeddings similar to ours. We share their motivation and idea but differ as (1) our method is not restricted to a predefined, row-by-row decoding order and uses a permutation-based training procedure aligned with the use of optimal, model-guided cell permutation during inference; (2) we assume the explicit prediction of the number of rows upfront (before the table decoding starts), instead of allowing the model to stop the generation process after any completed row. The advantage of this approach is discussed in Section~\ref{sec:permutation} and proven by a series of experiments reported in Section~\ref{sec:experiments}.}
%
%
%
%

\looseness=-1 The encoder-decoder model was previously used \textit{as is}, to infer lists and tuples separated with special characters \citep{powalski2021tilt,dwojak-etal-2020-dataset}. Similarly, \citet{DBLP:journals/corr/abs-2105-07510} experimented with the generation of more complex data types represented as XML, JSON, or Python’s string representation. One of the problems with these approaches is that the same output structure may be represented in various valid forms. Consequently, valid responses may be penalized during the training phase, expecting only one reference representation.
Due to the proposed grammar-constrained decoding process, we mitigate this problem without increasing the computational requirements that otherwise might result from the use of permutation-invariant loss functions. Moreover, in contrast to previous approaches, we do not rely on \textit{implicit} modeling of the formal structure of the output but opt for \textit{explicit} structure generation. {\color{black}The same remarks apply to the take of \citet{sage-etal-2020-end} who proposed a method for the extraction of XML-structured information based on a pointer-generation network. Additionally, in contrast to ours, their approach is limited to copying words from the input document and thus cannot perform normalization or return values that are not present in the text explicitly.}

\looseness=-1 Finally, a text-to-structure approach was recently taken by \citet{lu-etal-2021-text2event} for event extraction. The authors used trie-based constrained decoding with event schema injected as the decoder prompt. It resembles our approach to constrained table generation, though they rely on only one proper decoding order resulting from the assumed tree linearization strategy. Moreover, the authors found it challenging to train the structure generation model directly and thus trained it on simple event substructures first. In contrast, we can directly train the structure decoder, and our permutation-based method allows one to generate the structure \textit{flexibly}, in an arbitrary order dynamically guided by the decoding algorithm.

\textbf{Flexible generation.} Even though permutation-based training, which allows for output generation in any order, is of minor usability in the task of LM, it was validated by \citet{Stern2019InsertionTF} for machine translation. Accordingly, they proposed to equip a transformer with the insertion operation, realized by interpreting an additional number generated with the token as the position in the output sequence to which the insertion should be performed. This framework allows for the flexibility of the decoding process, understood as the possibility of stubbing the output sequence with tokens that the model recognizes with high confidence first and then gradually adding more details in the later iterations. In contrast, since the whole output sequence is passed through the decoder anyway, our one cell-decoding step is implemented by sampling all cells at once and then choosing the best-scored ones to be inserted at its location while disregarding others. In the ablation studies we evaluate how the number of cells inserted at once influence the decoding speed and quality, as higher values indicate more cells generated in parallel. 



\looseness=-1 \textbf{Permutation-based language modeling.} The effectiveness of the permutation-based language modeling objective was demonstrated by \citet{XLNET} who conditioned the BERT-like model to work with the AR objective. However, while the nature of the LM task allowed them to perturb the factorization order of the input sequence arbitrarily, our table-decoding problem requires additional constraints to account for the fact that each cell may consist of several tokens. Thus, the factorization order of blocks of tokens (representing cells) is permuted, while causal order is assumed within the cell.


\section{STable --- Text-to-Table Transformer Framework}


Serialized representation of the table permits to treat it as a text sequence, and hence, use text-centric methods to perform an autoregressive generation of the output sequence by employing a vanilla Transformer decoder. However, this approach does not exploit the two-dimensional structure of the table as it expands the answer sequentially and utilizes only uni-directional context.

Consequentially, two challenging problems arise. Firstly, how to approach the fact that some information in the table may depend on other cells (e.g., name and surname or the same tax rate for similar items on a receipt) while some may not be dependent (prices of different articles on the shopping list). In general, a model possesses flexibility with respect to this dependence-independence assumption when it can leverage dependencies during decoding but is not forced to do so in any specific order. Our idea is to solve this problem by delaying the generation of the most challenging and complex answers to later stages and conditioning them on the already generated answer.

\looseness=-1 The second issue that further complicates the problem is that the decoding must remain free of train-inference discrepancies. Generally, the train-inference alignment means that the state of the table at every step while decoding a particular example must also be possible to achieve in the training phase. Formulating the training that allows for flexible cell generation without providing any additional information remains a non-trivial problem.
We rise up to the challenge and demonstrate the solution below.

\label{sec:permutation}
\begin{figure}
    \centering
    \includegraphics[width=0.6\linewidth]{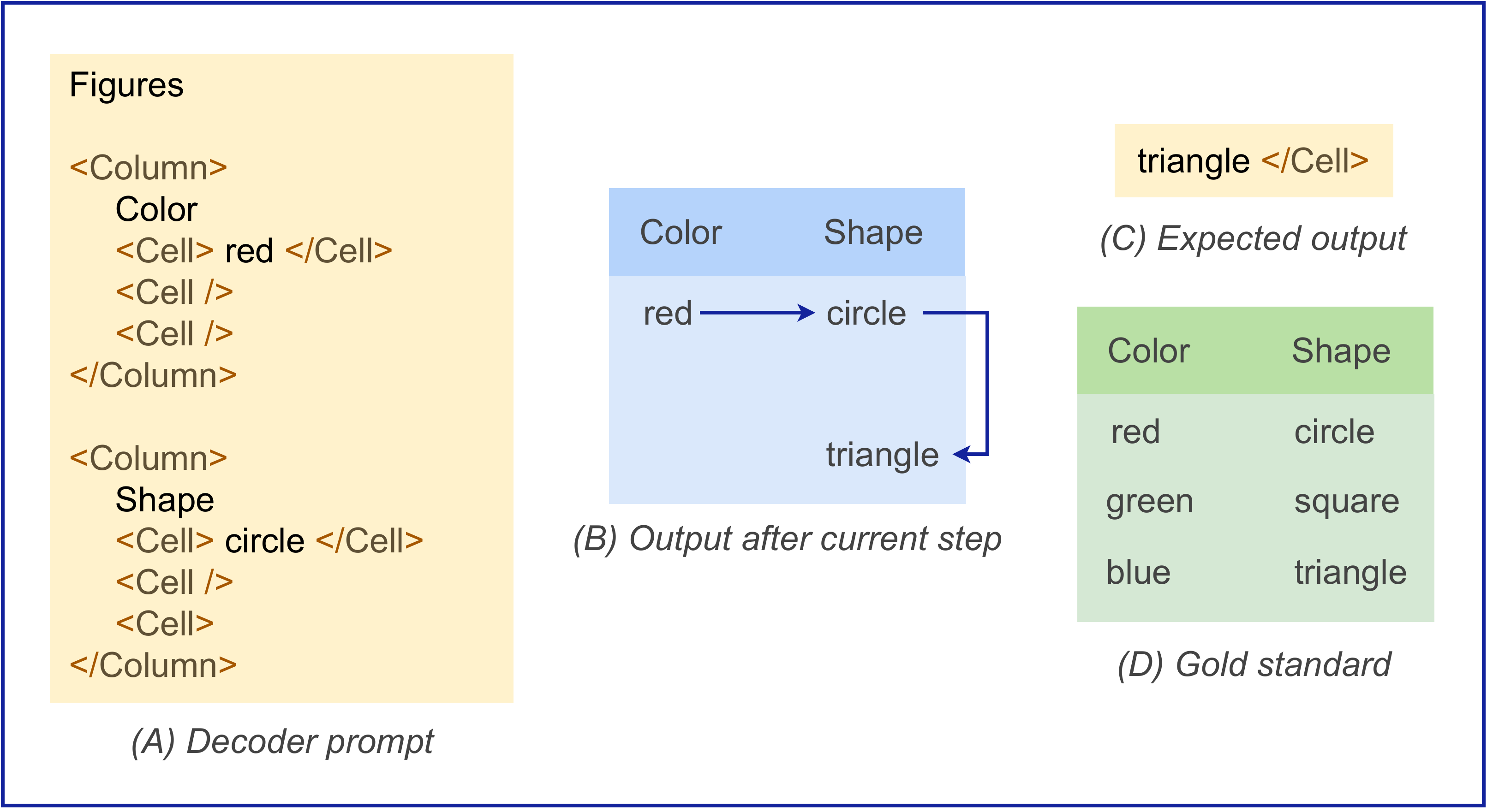}
    \caption{A training example starting after the second step in a sampled cell permutation. (A) Several columns contain empty cells (\texttt{<Cell />}, equivalent to \texttt{<Cell><Cell/>}), while the last cell within the sequence is not closed yet. (C) The latter's content is to be predicted by the model along with its closing tag. A successfully decoded cell will lead to the state visible in (B), i.e., the partially decoded gold standard table (D). Further decoding process within this sampled walk assumes the prediction of other empty cells.}
    \label{fig:training_example}
\end{figure}
\subsection{Decoding Invariant Under Cell Order}
Instead of generating the cell values in a top-down, left-to-right manner as previously seen in the literature \citep[e.g., ][]{DBLP:journals/corr/abs-2109-02707}, we perform the pretraining by maximizing the expected log-likelihood of the sequence of cell values over all possible prediction orders. 

More specifically, suppose that we are given a document containing a table with row labels $\mathbf{r} = (r_1,\dots, r_N),$\footnote{In practice, usually there are no row labels; however, in the decoder, the special tokens used for distinguishing rows take this role.} and column labels $\mathbf{c}=(c_1,\ldots,c_M)$, which we will collectively denote $\mathbf{h} = (\mathbf{r}, \mathbf{c})$. A linear ordering of the table cells can be represented with a bijection \[\sigma\colon\{1, 2,\dots, C\} \to \{1,\dots, N\} \times \{1,\dots, M\},\]
where $C=NM$ is the number of cells, so that $\sigma(n) = (i,j)$ are the row and column coordinates of the $n$-th cell in the ordering. Given such a $\sigma$ and cell values $\mathbf{v}=(v_{ij})_{i \leq N, j\leq M}$, we factorize the likelihood of $\mathbf{v}$ given $\mathbf{h}$ as
\begin{equation}
p_\theta(\mathbf{v} | \mathbf{h}) 
= \prod_{n=1}^{C} p_\theta\big(v_{\sigma(n)} \big|  (v_{\sigma(k)})_{k < n}, \mathbf{h}\big),
\end{equation}
and using this factorization, we maximize the expected log-likelihood
\begin{equation}
\frac{1}{C!} \sum_\sigma \sum_{n=1}^{C} \log p_\theta\big(v_{\sigma(n)} \big| (v_{\sigma(k)})_{k < n}, \mathbf{h}\big)
\end{equation}
over  $\theta$. The likelihoods $p_\theta\big(v_{\sigma(n)} \big|  (v_{\sigma(k)})_{k < n}, \mathbf{h}\big)$ themselves can be factorized according to the standard auto-regressive approach as
\red{
\begin{equation}
p_\theta\big(v_{\sigma(n)} \big| (v_{\sigma(k)})_{k < n}, \mathbf{h}\big) =
\prod_{t=1}^{\ell(v_{\sigma(n)})} p_\theta\big(v_{\sigma(n)}^t \big| (v_{\sigma(n)}^i)_{i < t} , (v_{\sigma(k)})_{k < n},  \mathbf{h}\big) 
\end{equation}
}\red{where $\ell(v_{\sigma(n)})$ is the length of $v_{\sigma(n)}$ represented as a sequence of tokens $(v_{\sigma(n)}^i)_{i\leq L}$. In practice, the expected log-likelihood is estimated by sampling bijections $\sigma$ at random.}

At its root, this approach is similar to the XLNet \citep{XLNET} permuted language modeling, though two significant distinctions are: (1) permutations are applied not to the encoder's input but to the decoder's, (2) the bidirectional contexts are assembled \textit{over} cells, while causal dependencies are modeled \textit{within} cells.

\subsection{Tabular Attention Bias}
\looseness=-1\red{We base our attention computation method on the relative bias idea popularized by the T5 model. Given a text consisting of $T$ tokens, in the vanilla T5 model, raw attention scores $\alpha_{ij}$ for tokens $i$ and $j$ (with $0 \leq i, j < T$) are modified by introducing a bias term: $\alpha'_{ij} = \alpha_{ij} + \beta_{ij}$ where $\beta_{ij} = W(i-j)$ is a trainable weight, depending on the relative sequential position of these tokens \citep{2020t5}.}

    

\looseness=-1 We modify the decoder's self-attention by extending it with two new bias terms, defined below. The \emph{tabular bias} $\tau_{ij}$ encodes the relative position of table cells in which the tokens lie, while the \emph{local sequential bias} $\lambda_{ij}$ corresponds to the relative sequential position of tokens belonging to the same cell.
    \begin{equation}
        \tau_{ij} = 
        \begin{cases}
            R(r_i-r_j) + C(c_i-c_j) & \text{if $r_j>0$} \\
            R_0 + C(c_i-c_j) & \text{if $r_j$ = 0}
        \end{cases},
        \qquad
        \lambda_{ij} = 
        \begin{cases}
            L(i-j) & \text{if $(c_i,r_i) = (c_j,r_j)$}\\
            0 & \text{otherwise}
        \end{cases}
    \end{equation}
    where $(c_{i}, r_i)$ are cell coordinates as given by its $1$-based column and row indices (with $0$ reserved for the header row/column), and $R(k)$, $C(k)$, $L(k)$ and $R_0$ are trainable weights. The special case with $r_j=0$ corresponds to the situation when the key/value token lies in the column header, in which case we want to use the same bias independent of the row of the query token, due to the different nature of the relation between two cells, and a cell and its column header.
    
    After these adjustments, the final attention score takes the form
    \begin{equation}
        \alpha'_{ij} = \alpha_{ij} + \beta_{ij} + \tau_{ij} + \lambda_{ij},
    \end{equation}
    where $\beta_{ij}$ is the bias term defined earlier.
\begin{figure}
    \centering
    \includegraphics[width=0.9\linewidth]{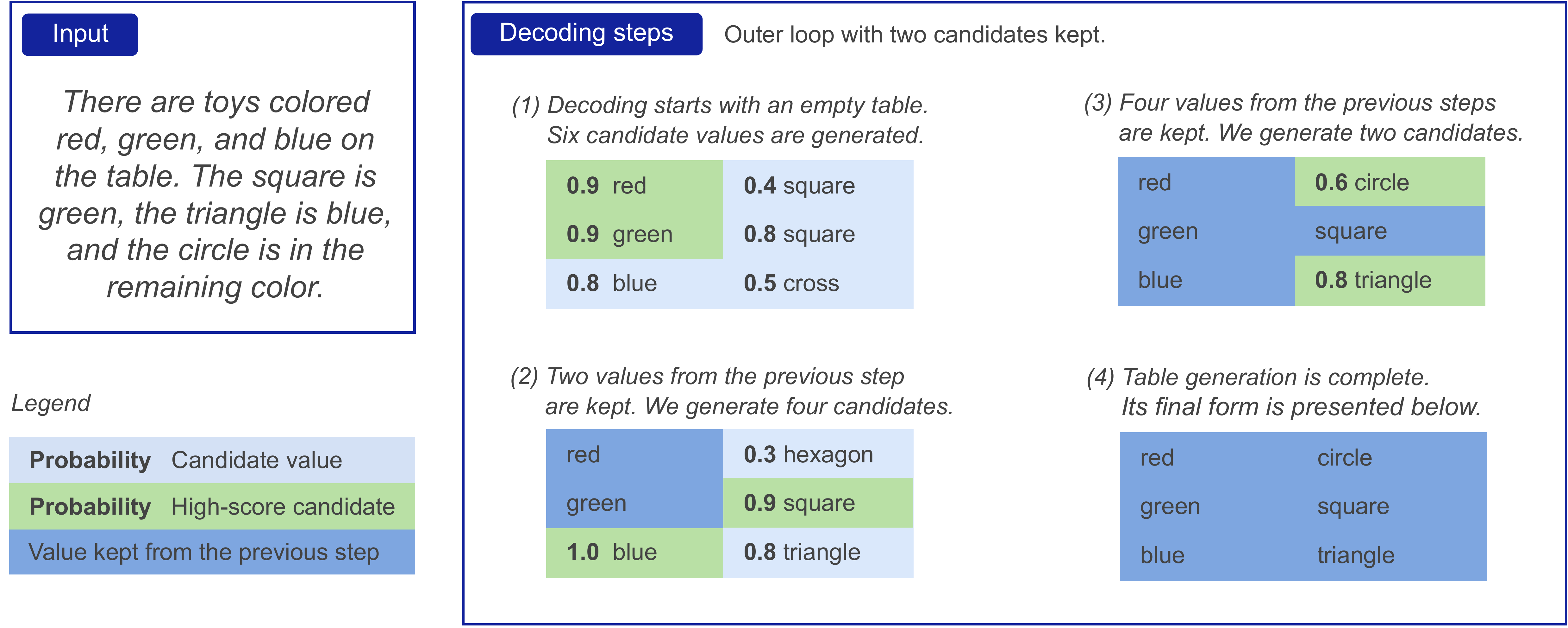}
    \caption{A possible progression of decoding a table given the text on the input. Since the probabilities guide the decoding order, the circle's color that was not explicitly stated in the text is determined at the last step.}
    \label{fig:decoding}
\end{figure}
{\color{black}Row and column bias we introduce is similar to the one used previously on the encoder side for comprehension of the two-dimensional structure of model input \citep{powalski2021tilt,garncarek2020lambert,xu-etal-2021-layoutlmv2,zayats-etal-2021-representations}. In contrast, we (1) apply it in the decoder, (2) do not use buckets with sizes growing logarithmically but have one bucket per row or column, and (3) have a dedicated bucket for the vertical relation between cell and its header.}

\subsection{Predicting Number of Groups}\label{sec:prediction}
Although the previous work of \citet{DBLP:journals/corr/abs-2109-02707} assumed the table is finalized when the appropriate special token explicitly appears in the output, our systematic study shows that the explicit prediction of the number of groups yields better results (see Section \ref{ablations} for comparison).
This explicit prediction is achieved with a linear layer that consumes the first input token's embedding to perform a prediction on the number of groups. During the training stage, the layer's output is scored against the known number of groups using MSE loss, while during the inference, it is used as a predictor declaring the number of groups to populate the template with.

\subsection{Inference with Model-Guided Cell Order} \label{sec:decoding}
While multiple decoding algorithms can be considered valid at test time, in light of the proposed training method we will focus on a simple greedy algorithm that does not introduce the train-inference discrepancy.

Since the model was trained assuming a permuted factorization of cell ordering, in expectation, the model learned to understand all possible variants of a partially-filled table and predict values for all empty cells. 
Because each step in the generation process implicates uncertainty that should be globally minimized, we propose to estimate the optimal table decoding algorithm by greedily finding the cell that minimizes this uncertainty at each step.

The decoding employs an outer loop that progresses cell-by-cell, an inner loop that generates each cell that is yet to render, and a selection heuristics that determine which cell, from all the finalized in the inner loop, should be added to the outer loop. The heuristic we use selects the cell containing the token with highest probability among all predicted (Figure~\ref{fig:decoding}). The detailed study of this and alternative selection criteria is presented in Table~\ref{tab:ablation}.

\looseness=-1 In the inner loop, each cell is decoded until the special token determining the end of cell generation is placed. As the inner loop generates each cell autoregressively and independently from other cells, the process can be treated as generating multiple concurrent threads of an answer and is well parallelizable. In the worst case, it takes as many steps as the number of tokens in the most extended cell.

After being selected by a heuristic, the cell from the inner loop is inserted into the outer loop, and made visible to all other cells, while the cells that were not selected are to be reset and continuously generated in the future steps until they are chosen by a heuristic  (please refer to Appendix~\ref{appendix:decoding} for the commented pseudocode and the algorithm details).

\section{Experiments}
\label{sec:experiments}

\begin{table}[]
\caption{Results on public and private datasets assuming task-specific metrics. The results of a sequence-to-sequence baseline that learns and generates tables as text are provided in the \textit{Linearized} column. Mean and STD over three runs. The $^\dagger$ symbol denotes our TILT training. Underline signifies our model is significantly better than baseline.}
\setlength{\tabcolsep}{5pt}
\renewcommand{\arraystretch}{0.85}
\label{tab:results}
\small
\centering
\begin{tabular}{llr|r|rl}
\toprule
Dataset & \multicolumn{2}{c}{\red{State-of-the-Art Reference}} & \multicolumn{1}{c}{\red{Linearized}} & \multicolumn{2}{c}{Our Model} \\
\midrule
PWC$^\bigstar$ & T5 2D \citep{due} & $26.8$ & \red{$27.8 \pm 1.0$} &  $ \boldsymbol{\underline{30.8}} \pm 0.5$ & T5 2D + STable \emoji{} \\ 
CORD & TILT \citep{powalski2021tilt} & $\boldsymbol{96.3}$ & \red{$92.4 \pm 0.7$} & $\underline{95.6} \pm 0.2$ & TILT$^\dagger$ + STable \emoji{} \\
\midrule
Rotowire \\
\quad Player & Text-to-Table \citep{DBLP:journals/corr/abs-2109-02707} & \red{$\boldsymbol{86.8}$} & \red{$84.5 \pm 0.7$} & $84.5 \pm 0.2$ & \multirow{2}{*}{T5 + STable \emoji{}}  \\
\quad Team & \red{\textit{(BART backbone)}} & \red{$\boldsymbol{86.3}$} & \red{$83.8 \pm 0.9$} & $\underline{84.7} \pm 0.2$ & \\
DWIE & KB-both \citep{verlinden-etal-2021-injecting} & $\boldsymbol{62.9}$ & \red{$60.2 \pm 1.5$} & $59.2 \pm 1.5$ & T5 + STable \emoji{} \\ 
\midrule
Recipe\ldots & \multirow{3}{*}{TILT$^\dagger$} & $71.9$ & \red{$60.1 \pm 0.3$} & $\boldsymbol{\underline{75.5}} \pm 1.6$ & \multirow{3}{*}{TILT$^\dagger$ + STable \emoji{}} \\ 
Payment\ldots & & $77.0$ & \red{$72.0 \pm 2.3$} & $\boldsymbol{\underline{79.1}} \pm 0.9$ & \\
Bank\ldots & & $61.1$ & \red{$58.7 \pm 4.9$} & $\boldsymbol{\underline{69.9}} \pm 4.8$ & \\
\bottomrule
\end{tabular}
\end{table}




\red{In addition to state-of-the-art reference and our results, we provide scores of the same backbone models (T5, T5 2D, and TILT) while assuming a table linearization strategy that follows the assumptions of \citet{DBLP:journals/corr/abs-2109-02707}'s baselines. Please refer to the Appendix~\ref{appendix:experiments} for details of evaluation metrics and training procedure.}

\textbf{Complex Information Extraction.} The problem of information extraction involving aggregated data types, where one may expect improvement within the document-to-table paradigm, is prevalent in business cases.
Nevertheless, the availability of public datasets here is limited to PWC$^\bigstar$ \citep{due,axcell} and CORD \citep{park2019cord}.

In the case of PWC$^\bigstar$, the goal is to determine model names, metrics, datasets, and performance, given the machine learning paper as an input. CORD assumes the extraction of line items from images of Indonesian receipts, among others. To determine the gain from our STable decoder, the experiments are conducted with state-of-the-art encoder-decoder models proposed for these datasets (T5 2D and TILT), assuming the same training procedure (\citet{due,powalski2021tilt}; see Appendix~\ref{appendix:experiments} for details).

Additionally, due to the sparsity of public benchmarks of this kind, we decided to provide results on three confidential datasets. They assume, respectively, (1) the extraction of payments' details from \textit{Payment Stubs}, (2) \textit{Recipe Composition} from documents provided by a multinational snack and beverage corporation, as well as (3) account balances from \textit{Bank Statements}.
These are covered in details in Appendix~\ref{appendix:business} and addressed by the TILT+STable model with vanilla TILT as a reference.

As summarized in Table~\ref{tab:results}, we outperformed state-of-the-art information extraction models on \red{several datasets. At the same time, the CORD where we underperform} was previously considered solved, e.g., \citet{powalski2021tilt} point that TILT's output and the reference differed insignificantly. We used it in the experiment as a safety check to determine whether the model can maintain almost-perfect scores after applying the STable decoder. Consequently, we omit it in the ablation studies.

The rest of the experiments were conducted assuming the vanilla T5 model \cite{2020t5} equipped with the STable decoder of our proposal.

\textbf{Joint Entity and Relation Extraction.}  To demonstrate the broad applicability of the model, we consider the problem of a joint entity and relation extraction on the example of the DWIE dataset \citep{dwie}. Here, the tuples consisting of entities and one of the sixty-five relation types are to be determined given a plain-text news article.
Despite not outperforming a multi-step state-of-the-art model, we achieved high scores and were the first to prove that the problem can be successfully approached end-to-end using an encoder-decoder framework. Here, the T5+STable's errors and issues reflect the very demanding assumptions of DWIE, where it is required to return \textit{object} and \textit{subject} in the longest form of appearance in the document.

\textbf{Reversed Table-to-Text.} Finally, following \citet{DBLP:journals/corr/abs-2109-02707} we evaluate our approach on the Rotowire table-to-text dataset in a reverse direction, i.e., generate tables from text \citep{wiseman-etal-2017-challenges}. Consequently, the complex tables reporting teams and player performance are generated given the game description.
\red{Results from Table~\ref{tab:results} show that our T5+STable model can deliver an improvement over the \textit{Linearized} T5 model on Rotowire Team.} 
\red{The fact that \textit{Linearized} BART from \citet{DBLP:journals/corr/abs-2109-02707} outperforms our \textit{Linearized} T5 baselines on Rotowire Team and Player datasets by $2.5$ and $2.1$ points, respectively, suggests that it has a better capacity as a backbone for this task.}
Several of the ablation studies from the next section were designed to shed light on this subject.

\red{The results of our model (Table~\ref{tab:results}) demonstrate a significant improvement over the simple sequence-to-sequence generation of tables linearized as sequences on three out of five public datasets. As expected, it yields better results in cases where there is a considerable interdependency between values in a row and no clear, known upfront name distinguishes it from other rows. Note that, e.g., in Rotowire, it suffices to correlate all statistics with team or player name, which is always inferred first due to the employed linearization strategy. The order of columns being decoded is a hyperparameter in the case of linearization. In contrast, the power of STable comes from learning it from the data itself.}

\section{Ablation Studies}\label{ablations}
Five ablation studies investigating the design choices' impact on performance were carried out. 
Models were trained three times with different random seeds on the Rotowire, DWIE, and PWC$^\bigstar$ datasets. To reduce the computational cost, we relied on \textit{base} variants of the models reported in Section~\ref{sec:experiments} -- please refer to Appendix~\ref{appendix:experiments} for detailed specifications and results. \red{While results on a single dataset can be considered noisy, aggregation over a diverse set of them helps diminish the randomness's impact and stabilize results on the new ones.}

\begin{table}[]
\renewcommand{\arraystretch}{0.85}
\caption{Results of studies (1), (2), (3), and (4). Modified models in relation to complete STable. See Appendix~\ref{appendix:experiments} for per-dataset results.}
\label{tab:ablation}
\small
\centering
\begin{tabular}{lrccl}
\toprule
Model & Average Score & Relative change & \\
\midrule
Complete STable (reference) & $62.9 \pm 1.0$ & --- \\
\midrule
Semi-templated expansion & $61.4 \pm 0.1$ & $-1.5$ & & (1) \\
\midrule
Fixed causal order & $60.0 \pm 0.4$ & $-2.9$ & & (2) \\
\midrule
Decoding constraint & & & & (3) \\
\quad Column-by-column & $62.4 \pm 0.6$ & $-0.5$ \\
\quad Row-by-row & $62.1 \pm 0.6$ & $-0.8$ \\
\quad L$\rightarrow$R and T$\rightarrow$B & $62.0 \pm 0.5$ & $-0.9$ \\
\quad No distant rows & $62.2 \pm 0.5$ & $-0.7$ \\
\midrule
Decision criteria (inner $\times$ outer) & & & & (4) \\
\quad min \hspace{3.5px} \quad max & $61.7 \pm 0.7$ & $-1.2$ \\
\quad mean \quad max & $62.7 \pm 0.7$ & $-0.2$ \\
\quad mean \quad min & $60.8 \pm 0.7$ & $-2.1$ \\
\quad min \hspace{3.5px} \quad min & $62.1 \pm 0.4$ & $-0.8$ \\
\quad max \hspace{2px} \quad min & $61.2 \pm 0.2$ & $-1.7$ \\
\bottomrule
\end{tabular}
\end{table}


\textbf{(1) Semi-templated Expansion.}
\looseness=-1 To compare our method of group prediction with a regression-free alternative, we allow the model to work in a semi-templated manner, where the template is infinite, and the decoding stops when the group with \textit{NULL}-only tokens is generated. For this method, we add such a group at the bottom of the table during the training to comply with the inference. The model performance is significantly below the STable reference, suggesting explicit group number prediction superiority.

\textbf{(2) Non-Permutative Training.}
\looseness=-1 To measure the importance of understanding the bidirectional contexts within the model, we abstain from permutation-based training in our study and choose the predefined factorization order. Here, a \textit{fixed causal order} model that reads tables from left to right and from top to bottom is evaluated. This alternative is in line with text-to-table approach of \citet{DBLP:journals/corr/abs-2109-02707}.
As shown in Table~\ref{tab:ablation}, the lack of permutative training we introduced in Section~\ref{sec:permutation} leads to significantly worse scores.

\textbf{(3) Constrained Cell Order.}
Whereas the permutation-based training allows for great flexibility, the questions posed here are whether limiting the model's ability to discover new cells can be of any value. Methods in this group assure either that the \textit{column-by-column} constrained model predicts the whole column before decoding a new one, the \textit{row-by-row} constrained model predicts the whole row before entering a new one, whereas \textit{L$\rightarrow$R and T$\rightarrow$B} is a combination of both (ensures row-by-row inference from left to right). The \textit{no distant rows} constraint forces the decoding to have empty cells only on the bottom of each column, thus avoiding skipping cells in the decoding while moving down.

\looseness=-1 As shown in Table~\ref{tab:ablation}, all but column-by-column constraint lead to a decreased scores. At the same time, the mentioned performs on par with STable's model-guided inference (Section~\ref{sec:decoding}), and both are better than the method with left-to-right decoding order. These results suggest that (1) our method does not require constraining the decoding order, (2) it seems to implicitly incorporate the column-by-column constraint, and (3) it is helpful to be elastic w.r.t. decoding order within the column.

\textbf{(4) Inner/Outer Loop Decision Criteria.}
The heuristic we test selects the cell in the outer loop based on the minimal or maximal inner score. Such inner score is calculated in three different ways: by taking the minimal, maximal, and mean of the token's logits score.
The results, presented in Table \ref{tab:ablation}, point to the lesser importance of choosing the inner scoring method, while the choice of the outer loop heuristics impacts results more significantly. The former is the desired behavior since the algorithm we proposed in Section~\ref{sec:decoding} is based on the assumption that it is beneficial to decode cells starting from those with the model's highest confidence. On the other hand, as there is a significant variance depending on the dataset chosen (see Appendix~\ref{appendix:experiments}), these and other inference parameters can be subject to cost-efficient, task-specific hyperparameter optimization.

\begin{figure}
    \centering
    \includegraphics[width=0.75\textwidth,trim={0.4cm 1.5cm 0.4cm  0.4cm},clip]{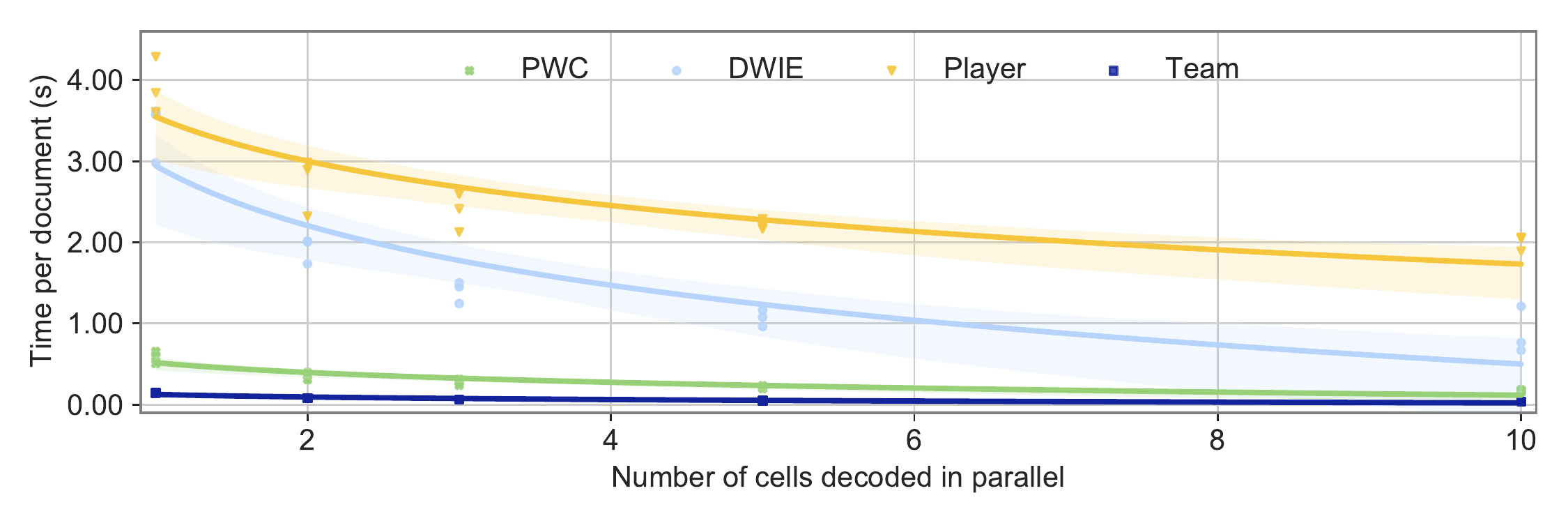}
    \includegraphics[width=0.75\textwidth,trim={0.4cm 0.5cm 0.4cm  0},clip]{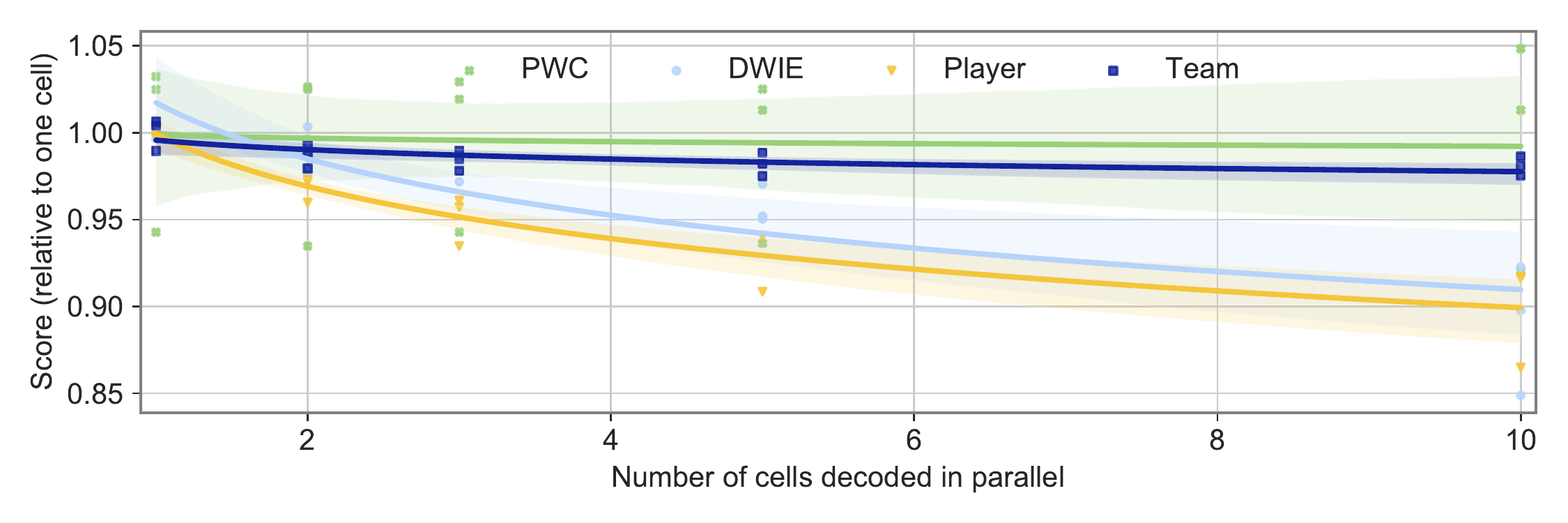}
    \caption{Results of decoding ablation (5). Three runs for 1, 2, 3, 5, and 10 cells decoded in parallel.}
    \label{fig:take}
\end{figure}

\textbf{(5) Parallelization of Cell Decoding.}
As outlined in Section~\ref{sec:decoding}, one may allow multiple candidates to be kept in each decoding step to shorten the inference time while expecting the performance to degrade to some extent. Results of experiments that follow this observation are presented in Figure~\ref{fig:take}.
One may notice that processing time varies across the considered datasets since it depends mainly on the input sequence length (ranging from $1k$ for Rotowire to $6k$ for PWC) and the sizes of the table to infer (we infer as many as $320$ cells for the Player table). 
Parallelization of cell decoding significantly reduces the total per-document processing time --- up to five times for DWIE in the conducted experiments. Interestingly, speed-up does not necessarily lead to a decrease in scores; e.g., in the case of the Team table, there is four times better processing time when ten cells are inferred at once, whereas the scores achieved by the model remain comparable. It can be attributed to the fact that there are almost no inter-cell dependencies and always only two rows to infer --- one for each team playing. Since the performance changes w.r.t. this parameter is heavily data-dependent, its value should be obtained experimentally for each dataset separately. Additionally, we argue that it is beneficial to use large values to speed up the train-time validation as it maintains a correlation with higher-scoring lower parameter values that can be employed during test-time inference.


\section{Summary}
\looseness=-1 We equipped the encoder-decoder models consuming text (T5, T5 2D) and documents (TILT) with the capabilities to generate tables in a data-dependent order. Firstly, an aligned training procedure based on permuting factorization order of cells was presented, and secondly, the parallelizable decoding process that fills the table with values in a flexible and unconstrained order was proposed. The important design choices for both contributions were motivated by an extensive ablation study. \red{The proposed STable framework demonstrates its high practical value by yielding state-of-the-art results on PWC$^\bigstar$ and outperforming linearized models on CORD and Rotowire Team datasets, as well as outperforming reference models on several confidential datasets.} \red{The highest gains due to the permutative training were accomplished on the PWC$^\bigstar$ dataset, where $4.0$ points ($26.8$ $\rightarrow$ $30.8$) amounts to $14.9\%$ relative improvement, while the $8.8$ point gain on Bank Statements ($61.1$ $\rightarrow$ $69.9$) exceeds $14.4\%$ relative improvement.}

\section*{Acknowledgments}

The Smart Growth Operational Programme supported this research under project no. POIR.01.01.01-00-1624/20 (\textit{Hyper-OCR --- an innovative solution for information extraction from scanned documents}).

\bibliography{bibliography}
\bibliographystyle{apalike}

\newpage
\appendix

\section{Table Decoding Algorithm}
\label{appendix:decoding}

\begin{algorithm}
    \caption{Table Decoding Algorithm of our proposal.} 
    \begin{algorithmic}[1]
    \Procedure{OuterLoop($k$)}{}
    \State $T
    \gets 0_{n,m,l}$ \Comment{$n \times m$ table with $l$ padding tokens per cell}
    \State $C 
    \gets 0_{n,m}$ \Comment{current cell status (decoded or not)}
    \While{$\Call{SUM}{C} < nm$} \Comment{while there is a cell to decode}
        \State $T', L \gets \Call{InnerLoop}{T, C}$ \Comment{create complete table candidate $T'$ and cell scores}
        \State $\mathcal{B} \gets \Call{OuterCriterion}{L}$ \Comment{sequence of coordinates sorted according to scores}
        \For{$c \gets 1, k$} \Comment{for $k$ best cells from T'}
            \State $i, j \gets \mathcal{B}_c$ \Comment{get coordinates}
            \State $T_{i,j} \gets T'_{i,j}$ \Comment{...copy values to table $T$ accordingly}
            \State $C_{i,j} \gets 1$ \Comment{...and mark the appropriate cell as already decoded}
        \EndFor
    \EndWhile
    \State \textbf{return} $T$
    \EndProcedure\\

    \Procedure{InnerLoop}{$T, C$}
    \State $L \gets 0_{n,m}$   \Comment{scores for each cell in $n \times m$ table}
    \State $T' \gets T$ \Comment{inner loop's table copy}
        \ParFor{$i \gets 1, n$} \Comment{for each table row}
            \ParFor{$j \gets 1, m$} \Comment{...and each table cell processed in parallel}
                \If{$C_{i,j} = 0$} \Comment{...if it was not decoded yet}
                    \State $s, t  \gets \Call{DecoderModel}{T, i, j}$ \Comment{produce cell tokens $t$ and their scores $s$}
                    \State $L_{i,j} \gets \Call{InnerCriterion}{s}$ \Comment{aggregate per-token scores into cell score}
                    \State $T'_{i, j} \gets t$ \Comment{update table copy}
                \EndIf
            \EndParFor
        \EndParFor
    \State \textbf{return} $(T', L)$
    \EndProcedure\\
    
    \Procedure{InnerCriterion}{$s$}
    \State \text{/* Any $\mathbb{R}^{n} \to \mathbb{R}$ function. STable assumes $max$, but we test other in the ablation studies. */}
    \EndProcedure\\

    \Procedure{OuterCriterion}{$L$}
    \State \text{/* Some  $\mathbb{R}^{m\times n}\to(\mathbb{N}\times\mathbb{N})^{mn}$
  function returning a permutation of indices of the input} \State \text{matrix $L$. STable assumes sort of matrix coordinates according to descending values of its}
  \State \text{elements, but we test other functions in the ablation studies. */}
    \EndProcedure\\
    
    
    \end{algorithmic}
    \label{alg:decoding}
\end{algorithm}

The algorithm presented above operates on the output of the encoder model and reuses the cached encoded representations that are considered to be a part of the \textsc{DecoderModel} for brevity. Another important characteristic of the \textsc{DecoderModel} introduced for conciseness of the pseudocode is that it produces all cell tokens and handles the sequential text decoding on its own.

The decoding employs an \textsc{OuterLoop}, parametrized by the $k$ parameter (denoting the parallelization of cell decoding) that progresses cell-by-cell, the \textsc{InnerLoop} function that generates each cell that is yet to render, and \textsc{OuterCriterion} --- a selection heuristics that determine which cell, from all the finalized in the inner loop, should be added to the outer loop. The \textsc{InnerCriterion} is a heuristic we utilize that selects the cell with the maximum probability for its tokens' predictions (Figure~\ref{fig:decoding}). 

In the \textsc{InnerLoop}, each cell is decoded until the special token determining the end of cell generation is placed. As the \textsc{InnerLoop} generates each cell autoregressively and independently from other cells, the process can be treated as generating multiple concurrent threads of an answer and is well parallelizable. In the worst case, it takes as many steps as the number of tokens in the most extended cell.

After the selection by the \textsc{OuterCriterion} heuristic, the cell from the inner loop is inserted into the outer loop, and made visible to all other cells, while the cells that were not selected are to be reset and continuously generated in the future steps until they are chosen by the \textsc{OuterCriterion} heuristics.

\section{Limitations}\label{sec:limitations}

\looseness=-1 The state-of-the-art performance of STable is its foremost advantage, while the constraining factors come from different aspects. Of them, the generated sequence's length seems to incure the most long-term cost during inference, while the increase in training time per example is a short-term obstacle.
The underlying issue is that the full table context negatively influences the computational cost of the attention on the decoder side. This however is also the case for the family of encoder-decoder models generating the whole table such as these proposed by \citet{DBLP:journals/corr/abs-2109-02707} or \citet{DBLP:journals/corr/abs-2105-07510}.
A possible solution here is a model with table context limited to the row and column a given table cell belongs to. Such a change would have a positive impact on the memory consumption in the decoder, as self-attention complexity decreases from $\mathcal{O}(NM)$ to $\mathcal{O}(N+M)$, where $N, M$ denotes the number of rows and columns respectively. The exploitation of this optimization is an interesting future direction.

Finally, our method by design does not generate the table header since we assume that the names of the datapoints to infer are given in advance. To tackle problems such as table structure recognition where the set of possible header values is not limited, one needs to slightly modify the proposed solution. However, we do not consider it a serious limitation as the required modification is relatively straightforward, and for the sake of completeness, we describe it in Appendix~\ref{appendix:structure_recognition}.

\section{Negative Result: Prevention of Column Order Leakage}

In the approach outlined in Section~\ref{sec:permutation}, the sequence of column labels $\mathbf{c}$, on which the likelihoods are conditioned, may leak additional unwanted information to the decoder. If the data in the document are indeed formatted as a table, and the order of labels in $\mathbf{c}$ matches the column order, the model might learn to extract cells by location, instead of using the actual semantics of the cell label. However, during inference, while we know which entities we want to extract from the document, we are not given the order in which they appear, which can be perceived as a serious train-inference discrepancy.

\begin{figure}
    \centering
    \includegraphics[width=1.0\linewidth]{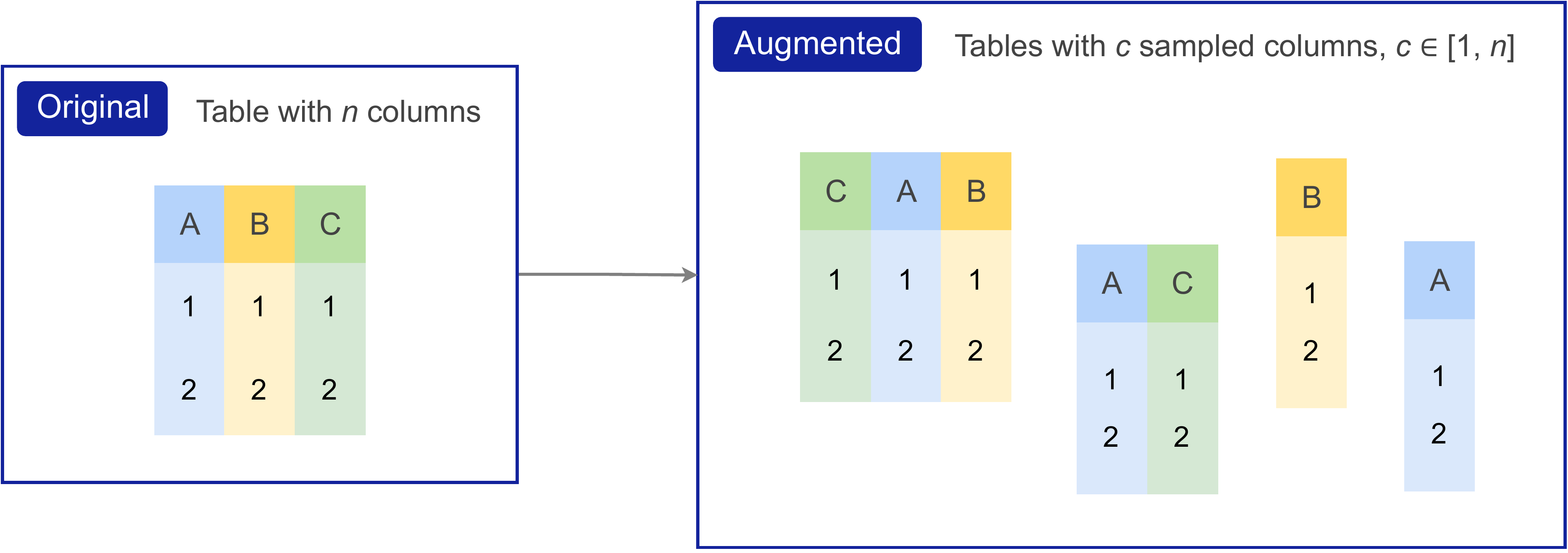}
    \caption{Change in training illustrated as augmentation of permuted sub-tables from the original table.}
    \label{fig:augmentation}
\end{figure}

To remedy this problem, we tried to further modify the training objective (See Figure~\ref{fig:augmentation}). Denote by $\CC$ the set of all non-empty sequences of distinct column labels. Instead of all the cells $\mathbf{v}$, we can predict only the cells $\mathbf{v}_\mathbf{c}$ corresponding to a sequence $\mathbf{c}\in\CC$ of columns, in the order defined by the order of columns in $\mathbf{c}$. The expected log-likelihood over all $\mathbf{c}\in\CC$ can be then expressed as
\begin{equation}
        \log p_\theta(\mathbf{v} | \mathbf{h}) = \frac{1}{\abs{\CC}} \sum_{\mathbf{c}\in\CC} \log p_\theta(\mathbf{v}_\mathbf{c} | \mathbf{r}, \mathbf{c}),
\end{equation}
where $p_\theta(\mathbf{v}_\mathbf{c} | \mathbf{r}, \mathbf{c})$ decomposes according to the discussion in Section~\ref{sec:permutation}.

In practice, we found it to have no relevant impact on the training process. It did not lead to significant changes in evaluation scores when used in the supervised pretraining stage or on a downstream task. Consequently, we abandoned the idea and did not use it for any of the models reported in the paper. This study helps us state that the model learns the semantics of the cell labels without a need for regularization.

\section{Details of Experiments and Ablation Studies\label{appendix:experiments}}

All models were trained three times with different random seeds. We relied on \textit{large} variants of the models for experiments in Table~\ref{tab:results}, and on \textit{base} variants for the ablation studies. These are analyzed in Table~\ref{tab:ablation} given the average results over Rotowire, PWC$^\bigstar$, and DWIE datasets (see Table~\ref{tab:ablation_detailed} for detailed scores).

\red{\textbf{Metrics.} We rely on the original metrics for all but the DWIE dataset, i.e., GROUP-ANLS for PWC$^\bigstar$, F1 for CORD, and non-header exact match cell F1 for Rotowire (other variants proposed by the authors are reported in Table~\ref{tab:rotowire}). Use of the original DWIE metric was not possible, as it assumes a step-by-step process. In contrast, we tackle the problem end-to-end, i.e., return \textit{(object, relation, subject)} tuples without detecting all entity mentions within the document and their locations. To ensure a fair comparison, we use the F1 score calculated on triples; that is, we require the model to return the exact match of the triple. Such a setup is very demanding for encoder-decoder models as the convention in DWIE is to require \textit{object} and \textit{subject} to be returned in the longest form of appearance in the document.}

\red{\textbf{Pretraining/adaptation.} Due to the switch to permutative training and the addition of the regression head, there is a significant change in the model objective. Consequently, we anticipated the necessity of the model adaptation phase. It consists of the pretraining stage equivalent to the one conducted by authors of the TILT model \citep{powalski2021tilt}, i.e., we use the same datasets and number of the model updates. The said stage is applied to both T5+STable, T5 2D+STable, and TILT+STable models.}

\begin{table}[]
\caption{Per-dataset results of studies (1), (2), (3), and (4). Modified models in relation to Complete STable.}
\label{tab:ablation_detailed}
\centering
\small
\begin{tabular}{lrrrccl}
\toprule
Model & RW Player & RW Team & PWC$^\bigstar$ & DWIE & \\
\midrule
Complete STable (reference) & $82.7 \pm 0.3$ & $84.1 \pm 0.7$ & $27.5 \pm 2.2$ & $56.0 \pm 1.4$ \\
\midrule
Semi-templated expansion & $80.4 \pm 0.5$ & $84.1 \pm 0.5$ & $25.0 \pm 0.8$ & $56.1 \pm 1.0$ & (1) \\
\midrule
Fixed causal order & $83.2 \pm 0.4$ & $84.3 \pm 0.3$ & $26.3 \pm 1.6$ & $46.5 \pm 0.5$ & (2) \\
\midrule
Decoding constraint & & & & & (3) \\
\quad Column-by-column & $82.5 \pm 0.4$ & $84.0 \pm 0.5$ & $28.4 \pm 1.5$ & $54.8 \pm 0.8$ \\
\quad Row-by-row & $80.2 \pm 0.4$ & $83.8 \pm 0.4$ & $27.6 \pm 1.6$ & $56.8 \pm 0.8$ \\
\quad L$\rightarrow$R and T$\rightarrow$B & $83.1 \pm 0.5$ & $84.1 \pm 0.7$ & $27.7 \pm 1.8$ & $53.2 \pm 0.5$ \\
\quad No distant rows & $82.7 \pm 0.5$ & $83.8 \pm 0.6$ & $28.1 \pm 1.0$ & $54.2 \pm 1.2$ \\
\midrule
Decision criteria (inner $\times$ outer) & & & & & (4) \\
\quad min \hspace{3.5px} \quad max & $81.9 \pm 0.4$ & $83.7 \pm 0.5$ & $26.5 \pm 2.0$ & $54.2 \pm 0.8$ \\
\quad mean \quad max & $83.0 \pm 0.3$ & $83.8 \pm 0.8$ & $27.8 \pm 1.4$ & $56.1 \pm 1.1$ \\
\quad mean \quad min & $81.2 \pm 1.1$ & $83.7 \pm 0.6$ & $26.4 \pm 1.9$ & $51.9 \pm 0.5$ \\
\quad min \hspace{3.5px} \quad min & $82.8 \pm 0.6$ & $83.8 \pm 0.5$ & $27.6 \pm 1.3$ & $54.0 \pm 0.5$ \\
\quad max \hspace{2px} \quad min & $82.3 \pm 0.3$ & $84.5 \pm 1.0$ & $20.7 \pm 1.6$ & $52.7 \pm 0.4$ \\
\bottomrule
\end{tabular}
\end{table}

\textbf{Hyperparameters.} We use task-independent hyperparameters that roughly follow these proposed by the authors of the T5 model for its finetuning, as during our initial experiments, they turned out to be a robust default (see Table~\ref{tab:hp}).

Maximal input sequence lengths were chosen in such a way a fair comparison with reference models was ensured. In particular, we use T5+2D's limit despite the fact one can achieve better results when consuming a more significant part of the input document. Similarly, the max number of updates follows the limit in reference models except for the DWIE dataset, where the state-of-the-art solution is based on the incomparable multi-step pipeline. See Table~\ref{tab:taskhp} for these task-specific details.


\textbf{Software and hardware.} All experiments and benchmarks were performed on DGX-A100 servers equipped with eight A100-SXM4-80GB GPUs that feature automatic mixed precision. Our models and references were implemented in PyTorch 1.8.0a0 \citep{NEURIPS2019_9015} with CUDA 11.4 and NVIDIA drivers 470.82.01.

\begin{table}
\caption{Task-independent hyperparameters used across all experiments.}
\label{tab:hp}
\small
\centering
\begin{tabular}{l|cccccc}
    \toprule
    Hparam & Dropout & Batch & Learning rate & Weight decay & Label smoothing & Optimizer \\
    Value & .1 & 64 & 1e-3 & 1e-5 & .1 & AdamW \\
    \bottomrule
\end{tabular}
\end{table}

\begin{table}
\caption{Task-dependent hyperparameters and training details. ($^*$) Length equal to the one consumed by the baseline model.}
\label{tab:taskhp}
\small
\centering
\begin{tabular}{lrrc}
    \toprule
    \multirow{2}{*}{Dataset} & \multicolumn{2}{c}{Max steps} & Max input \\
    & Ablation & Final & length \\
    \midrule
    PWC$^\bigstar$ & 500 & 1,000 & 6,144$^*$\hspace{-5px} \\
    Rotowire & 3,000 & 8,000 & 1,024 \\
    CORD & --- & 36,000 & 1,024 \\
    DWIE & 4,000 & 8,000 & 2,048 \\
    \midrule
    Recipe Composition & --- & 400 & 2600 \\
    Payment Stubs & --- \\
    Bank Statements & --- & 200 & 7000 \\
    \bottomrule
\end{tabular}
\end{table}

\begin{table}
\caption{Detailed results of experiments on reversed Rotowire dataset. See \citet{DBLP:journals/corr/abs-2109-02707} for metrics' specification.}.
\label{tab:rotowire}
\centering
\small
\begin{tabular}{lccccccccc}
    \toprule
    & \multicolumn{3}{c}{Row header F1} & \multicolumn{3}{c}{Column header F1} & \multicolumn{3}{c}{Non-header F1}\\
    & Exact & Chrf & BERT & Exact & Chrf & BERT & Exact & Chrf & BERT \\
    \midrule
    Team & 94.9 & 95.2 & 97.8 & 88.9 & 85.8 & 88.7 & 84.7 & 85.6 & 90.3 \\
    Player & 93.5 & 95.3 & 95.1 & 88.1 & 91.2 & 94.5 & 84.5 & 86.8 & 90.4 \\
    \bottomrule
\end{tabular}
\end{table}

\section{Business Datasets}\label{appendix:business}

\begin{table}[]
\caption{Summary of the confidential datasets.}
\small
\centering
\begin{tabular}{lrrr}
\toprule
  & \multicolumn{1}{c}{Recipe Composition} & \multicolumn{1}{c}{Payment Stubs} & \multicolumn{1}{c}{Bank Statements} \\
\midrule
train documents & $119$ & $80$ & $111$ \\
val documents & $16$ & $10$ & $10$ \\
test documents & $30$ & $20$ & $10$ \\
\midrule
avg doc len (words) & $0.6$k & $0.3$k & $1.3$k \\
max doc len (words) & $1.6$k & $2$k & $4,9$k \\
avg doc len (characters) & $3.3$k & $2$k & $8.3$k \\
max doc len (characters) & $10$k & $14.2$k & $37.9$k \\
\midrule
properties total & $64$ & $11$ & $10$ \\
properties in tables (tables columns) & $64$ & $4$ & $4$ \\
properties outside of tables & $0$ & $7$ & $6$ \\
mean number of table rows & $12$ & $5$ & $2$ \\
max number of rows & $60$ & $15$ & $5$ \\
\midrule
mean length of cell (characters) & $12$ & $8$ & $9$ \\
max length of cell (characters) & $308$ & $44$ & $36$ \\

\bottomrule \\
\end{tabular}
\label{tab:private_datasets}
\end{table}

Due to the sparsity of public benchmarks for complex information extraction, we decided to provide results on three confidential datasets. They assume, respectively, (1) the extraction of payments' details from \textit{Payment Stubs}, (2) \textit{Recipe Composition} from documents provided by multinational snack and beverage corporation, as well as (3) account balances from \textit{Bank Statements}. Their details are covered in the present section and Table~\ref{tab:private_datasets}.

\textbf{Recipe Composition.} The problem faced is extracting proprieties of food ingredients from confidential food manufacturer's documentation. This dataset contains 165 annotated fragments from 55 documents, three pieces for each document, with annotations sourced from the corporation's CRM system.

For each file, there are five tables to be extracted. The first one describes the ingredient's physical and chemical parameters (i.e., parameter name, testing method, range of allowed values, unit of measurement, and testing method details). The second one describes sub-components of the ingredient (i.e., its quantity, name, allergens, ingredient function, and country of origin). The third table informs about the presence of allergens (e.g., their names and binary information about their presence). The last two tables contain a quantity of the allergens (e.g., names and their qualities) as sub-components and caused by contamination retrospectively.

The first table needs to be extracted from the first document fragment, the second table -- from the second fragment, and the three last tables -- from the third document fragment. Input documents feature tables and fulfilled forms, where properties are presented in the form of text or check-boxes.

The analysis of expected outputs shows a high level of variability concerning the factors of table length (1 to 60 rows) and answer type (either a binary value, number, complex chemical name, or a more extended description).

\textbf{Payment Stubs.} The second of our private datasets consists of 110 American payment stubs, i.e., documents obtained by an employee regarding the salary received.

We aim to extract employee and employer names, dates, and payment tables, where each row consists of payment type, hours worked, and payment amount. Since documents come from different companies, their layouts differ significantly. 

Due to the straightforward form of information to be extracted, a single annotator annotated each document. We state these were annotated ethically by our paid co-workers.

\textbf{Bank Statements.} The last dataset consists of 131 annotated bank statements. The goal here is to extract bank and customer name, date of issue, and table of account balances (e.g., account number, balance at the beginning of the period, and balance at the end).

Data to be comprehended is partially presented in the document's header and partially in multiple forms (each for one account).

Similar to the Payment Stubs dataset, documents here were issued by different banks and represent a broad spectrum of layouts. The annotation process was the same as for the Payment Stubs dataset.

\section{Adaptation to Table Structure Recognition Task}\label{appendix:structure_recognition}
To adjust the proposed method to be applicable to the task of Table Structure Recognition, one must understand the differences in framing the problem between the tasks here.

Table Structure Recognition or Table Extraction aims to generate headers and the table content based on the document with the table provided explicitly. 
STable described in the main part of this paper can generate the table given any text and its position on pages. This capacity generalizes well to any input, including when the table is provided on the input. The difference is that the output form in STable assumes the headers are known upfront, while for Table Structure Recognition, inferring them is a part of the task. STable can achieve such capabilities to solve the Table Structure Recognition task by (1) adding a linear layer to predict the number of columns, (2) treating headers as the values to be inferred in the first row, (3) using dummy names of the columns, e.g., "first column," "second column," and  (4) increasing the predicted number of rows by $1$.

In this setup, the model will predict the number of columns and the number of rows, while the first row will represent the values of header names. The dummy headers will have to be removed during postprocessing, and the values in the first row should be treated as valid headers.

\section{\red{Sample Input-Output Pairs}}

\begin{figure}
    \centering
    \includegraphics[width=\linewidth]{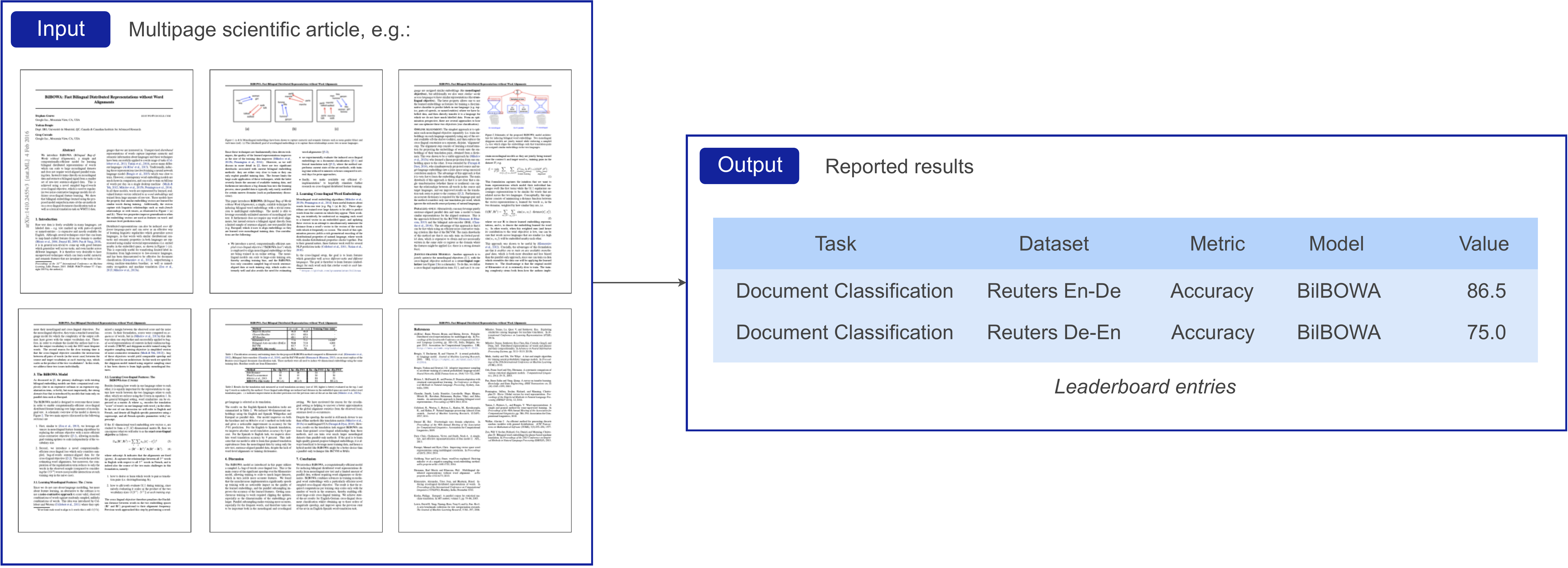}
    \red{\caption{An example from PWC$^\bigstar$ dataset considered in the document-to-table paradigm.}}\vspace{5mm}
    \label{fig:pwc_example}
\end{figure}

\red{\textbf{PWC$^\bigstar$} \citep{due}. Input in the PWC$^\bigstar$ consists of born-digital, multipage PDF files containing an article from the machine learning field. The expected output is a list of tuples describing achieved results on arbitrary datasets (see Figure~\ref{fig:pwc_example}).}

\begin{figure}
    \centering
    \includegraphics[width=0.95\linewidth]{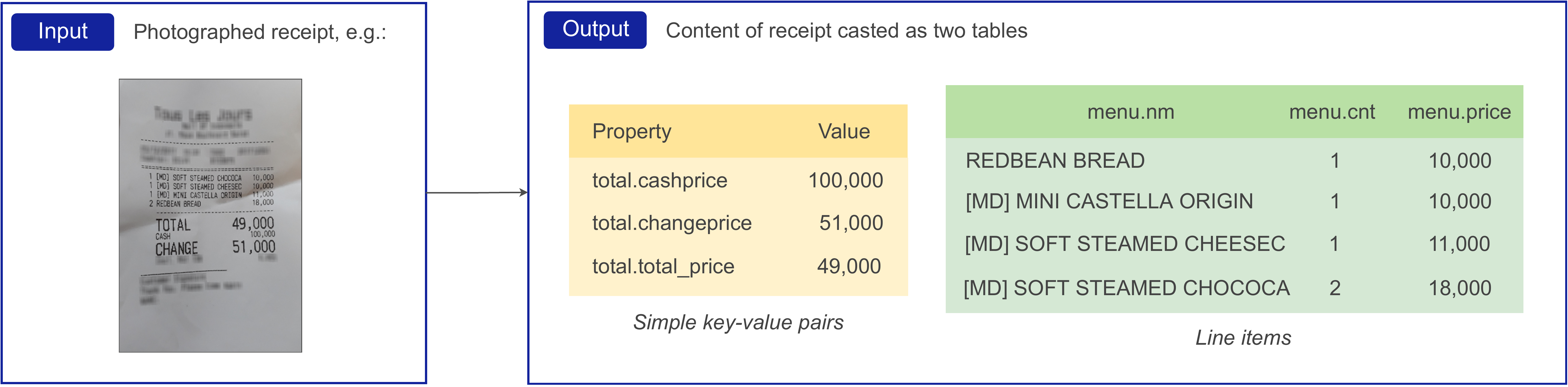}
    \red{\caption{Sample document from CORD dataset and its expected output as interpreted in our approach.}}
    \label{fig:cord_example}
\end{figure}

\red{\textbf{CORD} \citep{park2019cord}. Input in the dataset is a single scanned or photographed receipt. From our point of view, the output here is twofold --- there are simple data points that can be considered key-value pairs and data points that take the structured form of line items. We approach the problem as the generation of two tables from the document --- one for each data kind (see Figure~\ref{fig:cord_example}).}

\begin{figure}
    \centering
    \includegraphics[width=0.85\linewidth]{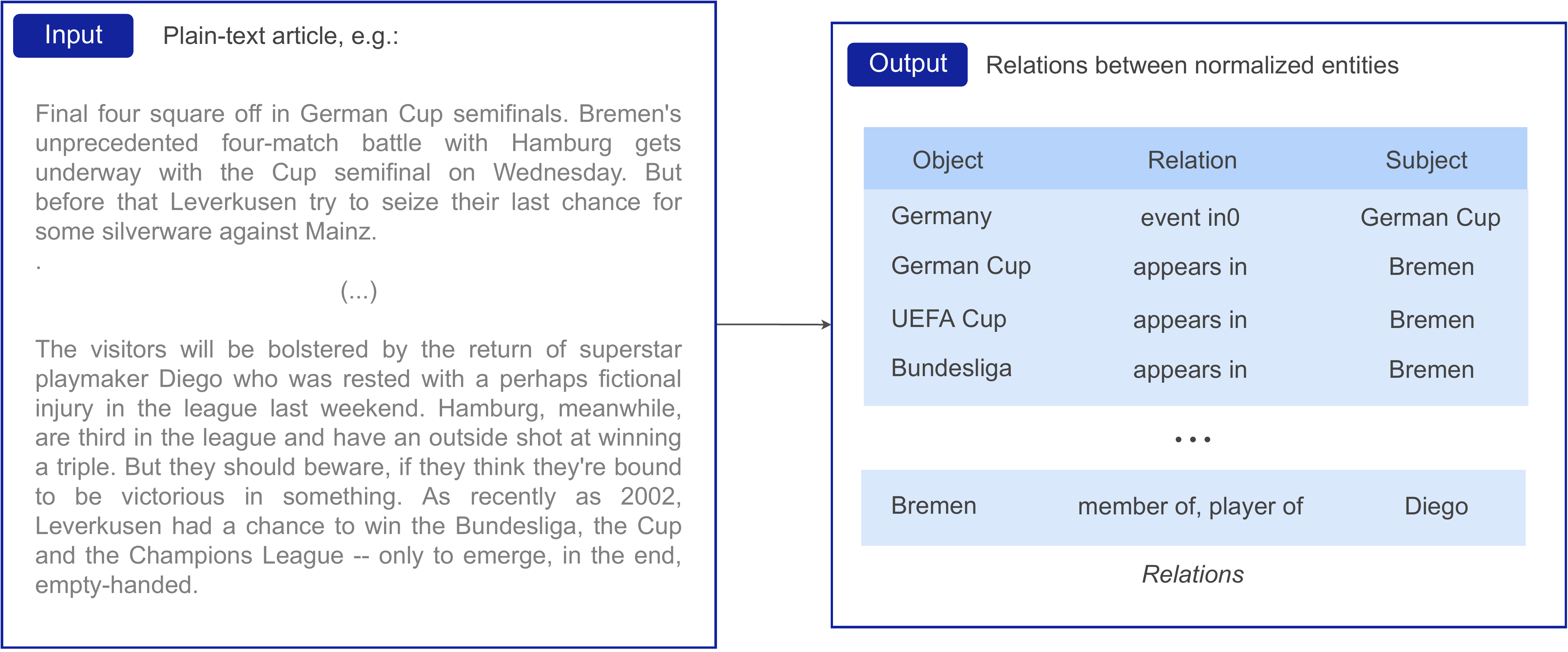}
    \red{\caption{Sample input-output pair from the DWIE dataset. The table was shortened and consisted of 29 rows in our approach. Suppose multiple relations appear in the same direction between the pair of object-subject. In that case, we predict a list of them in a single cell, reducing the number of rows generated (see the example of the Bremen-Diego pair).}}
    \label{fig:dwie_example}
\end{figure}

\begin{figure}
    \centering
    \includegraphics[width=0.8\linewidth]{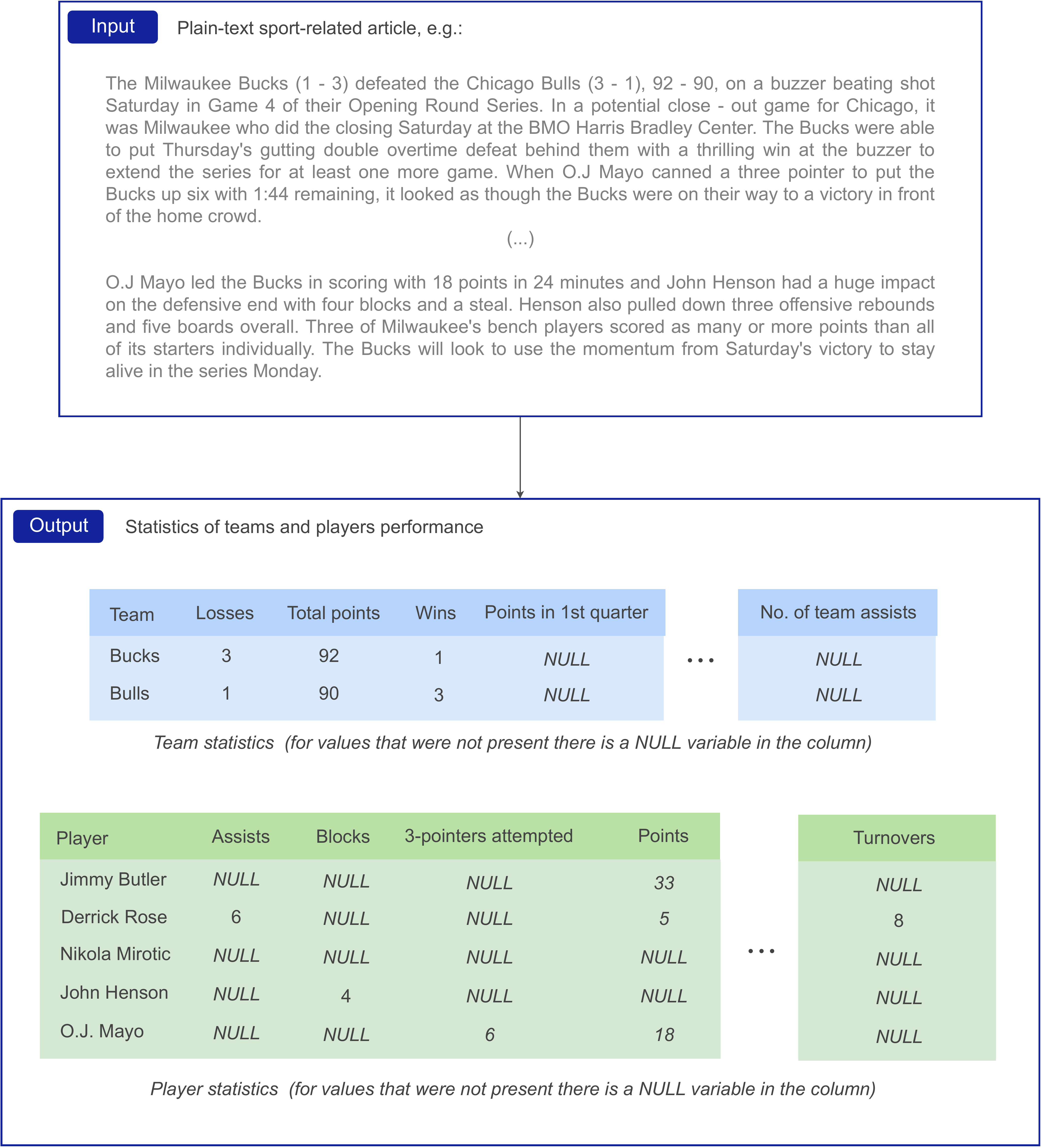}
    \red{\caption{Input-output example from the reversed Rotowire dataset. We present shortened forms of tables than in real have 13 columns for Team and 20 columns for Player tables. Note that there is a \textit{NULL} value in the column for values not present in the input text.}}
    \label{fig:rotowire_example}
\end{figure}

\red{\textbf{Reversed Rotowire} \citep{DBLP:journals/corr/abs-2109-02707}. Input in the reversed Rotowire dataset, as reformulated by \cite{DBLP:journals/corr/abs-2109-02707}, is a plain-text sport news article. The task is to generate tables with team and player statistics. The number of rows in the \textit{Team} table is from zero (if no team is mentioned in the text) to two, whereas the number of rows in the \textit{Player} is highly variable and content-dependent. Figure~\ref{fig:rotowire_example} present sample pair of document and tables to generate.}

\red{\textbf{DWIE} \citep{dwie}. Input in the dataset is a plain-text article. The final goal is to extract the normed object, relation, and subject triples (though the original formulation assumes several intermediate stages). Triples are always complete (i.e., there are no NULL values, as we understand them (see Figure~\ref{fig:dwie_example} for an example).}

\vspace*{\fill}
\centering\includegraphics[width=30px]{images/emoji_u1f434.pdf}\\\vspace{5px}
\textit{The horse face emoji we feature is a part of Noto Emoji distributed under the Apache License 2.0. Copyright by Google Inc. No animals were harmed in the making of this article.}
\end{document}